\documentclass[letterpaper, 10 pt, conference]{ieeeconf}  

\IEEEoverridecommandlockouts                              

\usepackage{graphics} 
\usepackage{epsfig} 
\usepackage{mathptmx} 
\usepackage{times} 
\usepackage{amsmath} 
\usepackage{amssymb}  
\usepackage{subfigure}
\usepackage{pifont}
\usepackage{cite}

\usepackage{color}
\usepackage{etoolbox}

\title{\LARGE \bf
Aerial-PASS: Panoramic Annular Scene Segmentation in Drone Videos
}

\author{Lei Sun$^{1}$, Jia Wang$^{1}$, Kailun Yang$^{2}$, Kaikai Wu$^{1}$, Xiangdong Zhou$^{1}$, Kaiwei Wang$^{1}$ and Jian Bai$^{1}$
\thanks{This research was granted from ZJU-Sunny Photonics Innovation Center (No. 2020-03). This research was also funded in part through the AccessibleMaps project by the Federal Ministry of Labor and Social Affairs (BMAS) under the Grant No. 01KM151112. This research was also supported in part by Hangzhou HuanJun Technology Company Ltd. and Hangzhou SurImage Technology Company Ltd.}
\thanks{$^{1}$Authors are with State Key Laboratory of Optical Instrumentation, Zhejiang University, China {\tt \{wangkaiwei, bai\}@zju.edu.cn}}%
\thanks{$^{2}$The author is with Institute for Anthropomatics and Robotics, Karlsruhe Institute of Technology, Germany}%
\thanks{Our dataset will be made publicly available at: http://wangkaiwei.org}
}
\begin{document}

\maketitle
\thispagestyle{empty}
\pagestyle{empty}

\begin{abstract}

Aerial pixel-wise scene perception of the surrounding environment is an important task for UAVs (Unmanned Aerial Vehicles).
Previous research works mainly adopt conventional pinhole cameras or fisheye cameras as the imaging device. However, these imaging systems cannot achieve large Field of View (FoV), small size, and lightweight at the same time.
To this end, we design a UAV system with a Panoramic Annular Lens (PAL), which has the characteristics of small size, low weight, and a $360^\circ$ annular FoV.
A lightweight panoramic annular semantic segmentation neural network model is designed to achieve high-accuracy and real-time scene parsing.
In addition, we present the first drone-perspective panoramic scene segmentation dataset Aerial-PASS, with annotated labels of track, field, and others.
A comprehensive variety of experiments shows that the designed system performs satisfactorily in aerial panoramic scene parsing.
In particular, our proposed model strikes an excellent trade-off between segmentation performance and inference speed suitable, validated on both public street-scene and our established aerial-scene datasets.

\end{abstract}

\section{Introduction}

In the last years, UAV (Unmanned Aerial Vehicle) systems have become relevant for applications in military recognition, civil engineering, environmental surveillance, rice paddy remote sensing, and spraying, etc.~\cite{adams2011survey,d2012unmanned}.
Compared with classic aerial photography and ground photography, the UAV system is more flexible, small in size, low-cost, and suitable for a wider range of application scenarios. Environment perception algorithms based on UAV system needs to be light and efficient enough for the application in mobile computing devices like portable embedded GPU processors.

\begin{figure}[!t]
    \centerline{\includegraphics[width = 0.45\textwidth]{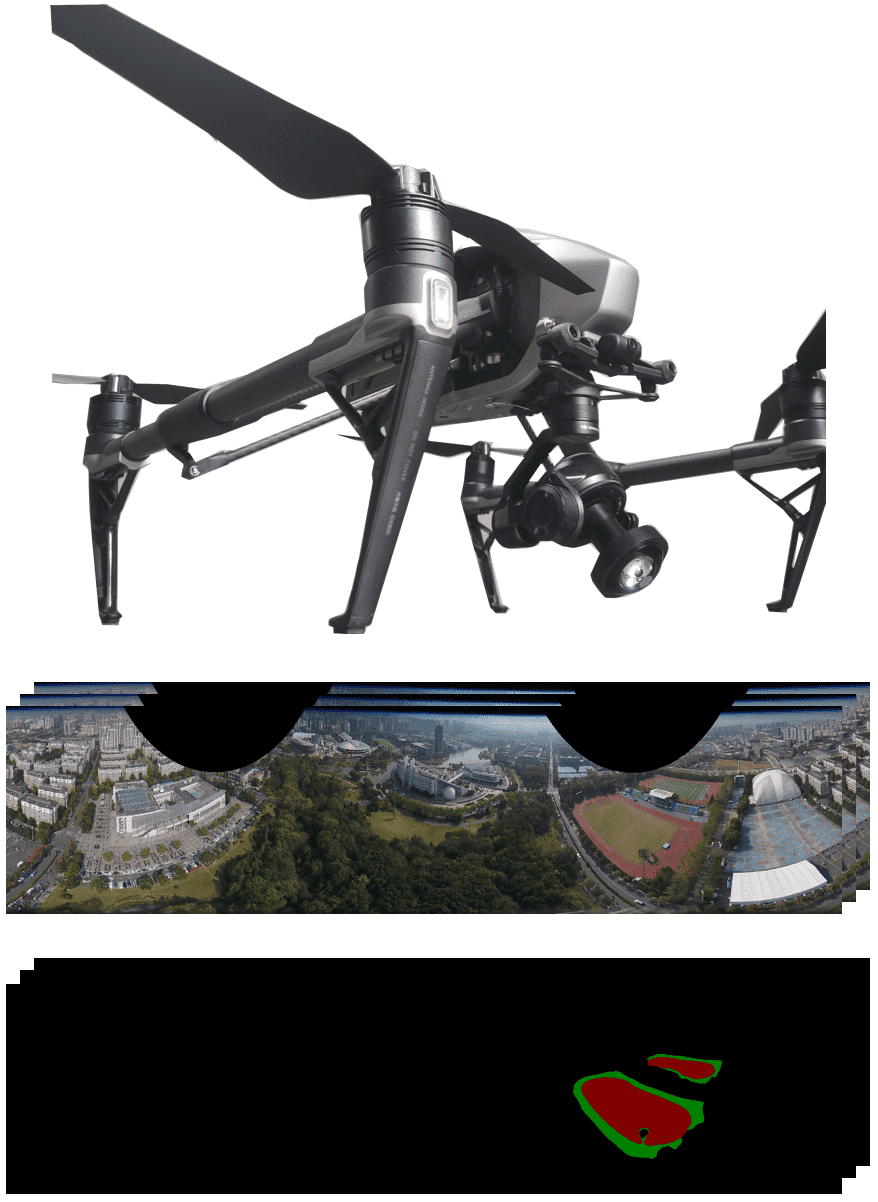}}
    \caption{Overview of our Aerial-PASS system. Panoramic images captured from the UAV with a PAL camera are unfolded, and with the designed semantic segmentation model, field and track classes are predicted at the pixel level.}
    \label{fig:overview}
\end{figure}

However, most optical lens of the UAV systems have a small field of view, and often rely on a complex servo structure to control the pitch of the lens, and post-stitch the collected images to obtain a $360^\circ$ panoramic image~\cite{d2012unmanned}.
The expansion of the field of view is essential for real-time monitoring with UAVs.
In traditional methods, the control system of UAV is usually very complicated.
Since the drone takes images during flight, the post-stitching algorithm is highly computation-demanding, and the images have parallax and exposure differences, which renders the reliability of image stitching rather low.
To address this problem,
we have designed a lightweight Panoramic Annular Lens (PAL) especially suitable for UAV systems.
The system does not require a complicated servo system to control the attitude of the lens. The optical axis of the lens is placed vertically on the ground, and the cylindrical field of view can be horizontally upward by $10^\circ$ and downward by $60^\circ$ (Fig.~\ref{fig:overview}). 

To support fast on-board remote sensing, we further propose a lightweight real-time semantic segmentation network for panoramic image segmentation.
The network has an efficient up-sampling module and a multi-scale pooling module for learning objects of different scales in the panoramic images. To facilitate credible evaluation, we collect with our PAL-UAV system and present the first drone-perspective panoramic scene segmentation benchmark Aerial-PASS with annotated labels of critical field sensing categories.
We find a superior balance between accuracy and inference speed for the network towards efficient aerial image segmentation and it also achieves the state-of-the-art real-time segmentation performance on the popular \emph{Cityscapes} dataset~\cite{Cordts2016Cityscapes}. 

The contributions of this paper are summarized as follows:

\begin{itemize}

\item The designed PAL lens has the advantages of $4K$ high resolution, large field of view, miniaturization design, and real-time imaging capabilities, etc. It can be applied to UAV survey and identification, autonomous driving, security monitoring, and other fields.
\item An efficient real-time semantic segmentation network is proposed for panoramic images and it achieves a state-of-the-art accuracy-speed trade-off on \emph{Cityscapes}. A set of comparison experiments is conducted on the panoramic images collected by our PAL-UAV system. 
\item An annotated Aerial Panoramic dataset is presented for the first time, which is conducive to the rapid and robust segmentation of target objects in a large field of view from UAV perspectives. Particularly, this work focuses on pixel-wise segmentation of track and field objects.
\end{itemize}
\section{Related Works}

\subsection{Panoramic Annular Imaging}

The optical lens in the drone provides an indispensable visual aid for target recognition and detection. A single large field of view lens can bring a wider field of view to the drone.
The current ultra wide-angle optical systems include fisheye lenses~\cite{fisheye}, catadioptric optical systems~\cite{catadioptric}, and Panoramic Annular Lens (PAL) imaging systems~\cite{panoramic_lens}.
The structure of the PAL is simpler, smaller in size, easy to carry, and better in imaging quality. These advantages make the PAL become the research focus of large field of view optical systems.

The PAL imaging system was proposed by Greguss in 1986~\cite{greguss1986panoramic}. In 1994, Powell designed the infrared band large field of view PAL imaging system, and showed that the spherical surface type can be used to obtain better imaging quality~\cite{powell}.
A cluster of scientific research teams has also made great research progress, designing PAL optical systems with longer focal length~\cite{niu}, larger field of view~\cite{large_field_of_view}, higher resolution~\cite{wang2019design}, and better imaging quality~\cite{high-perfoemance}.
The PAL designed in this work is small in size, easy to carry, and it has a field of view of $(30^\circ-100^\circ)\times360^\circ$
and a $4K$ high resolution, which can realize real-time large field of view high-resolution single sensor imaging. Thereby, it is perfectly suitable for UAV survey, unmanned driving, security monitoring, and other application fields~\cite{fang2020cfvl,chen2021panoramic}.

\subsection{Panoramic Scene Segmentation}

Beginning with the success of Fully Convolutional Networks (FCNs)~\cite{fcn}, semantic segmentation can be achieved in an end-to-end fashion. Subsequent composite architectures like PSPNet~\cite{pspnet} and DeepLab~\cite{deeplab} have attained remarkable parsing performance.
Yet, due to the use of large backbones like ResNet-101~\cite{resnet}, top-accuracy networks come with prohibitively high computational complexity, which are not suitable for mobile applications such as driving- and aerial-scene segmentation in autonomous vehicles or drone videos. 
Plenty of light-weight networks emerge such as SwiftNet~\cite{swiftnet}, AttaNet~\cite{attanet}, and DDRNet~\cite{ddrnet}, both seeking a fast and precise segmentation.
In our previous works, we have leveraged efficient networks for road scene parsing applications like nighttime scene segmentation~\cite{see_clearer_at_night} and unexpected obstacle detection~\cite{rfnet}.

Driving and aerial scene segmentation clearly benefit from expanded FoV, and standard semantic segmentation on pinhole images has been extended to operate on fisheye images~\cite{universal}, omnidirectional images~\cite{omniscape}, and panoramic annular images~\cite{pass}.
The latest progress include omni-supervised learning~\cite{ooss} and omni-range context modeling~\cite{omnirange} to promote efficient $360^\circ$ driving scene understanding.
In particular, Yang~\textit{et al.}~\cite{pass} proposed a generic framework by using a panoramic annular lens system, which intertwines a network adaptation method by re-using models learned on standard semantic segmentation datasets.
Compared to driving scene segmentation, aerial image segmentation is still predominantly based on pinhole images~\cite{agriculture_vision,relation_augmented,aerial_lanenet,pointflow}.
In this paper, we lift $360^\circ$ panoramic segmentation to drone videos and propose an Aerial-PASS system to explore the superiority of the ultra-wide FoV for security applications.

\section{Proposed System}

The proposed system consists of the hardware part and the algorithm part.
In the hardware part, we have equipped the UAV with our designed PAL camera and use it to collect panoramic image data.
To efficiently parse panoramic images, we have designed a lightweight semantic segmentation model with a novel multi-scale receptive field pyramid pooling module for learning a robust feature representation required for $360^\circ$ image segmentation.

\subsection{UAV with PAL Camera}

The PAL system designed and used in this work has the characteristics of large field of view and lightweight structure, which can realize real-time, single-sensor imaging on drones.
The PAL system follows the imaging principle of Flat Cylindrical Perspective~(FCP), which can reflect light by the panoramic block from the lateral field of view around the optical axis $360^\circ$, and then enter the subsequent lens group and image on the two-dimensional image surface. The field of view of the PAL system is generally greater than $180^\circ$, which is no longer applicable to the classic principle of object-image similarity. At this point, we introduce negative distortion in the optical design to control the height of the image surface, and the commonly used characterization method is F-Theta distortion.

\begin{figure}[!t]
    \centerline{\includegraphics[width = 0.45\textwidth]{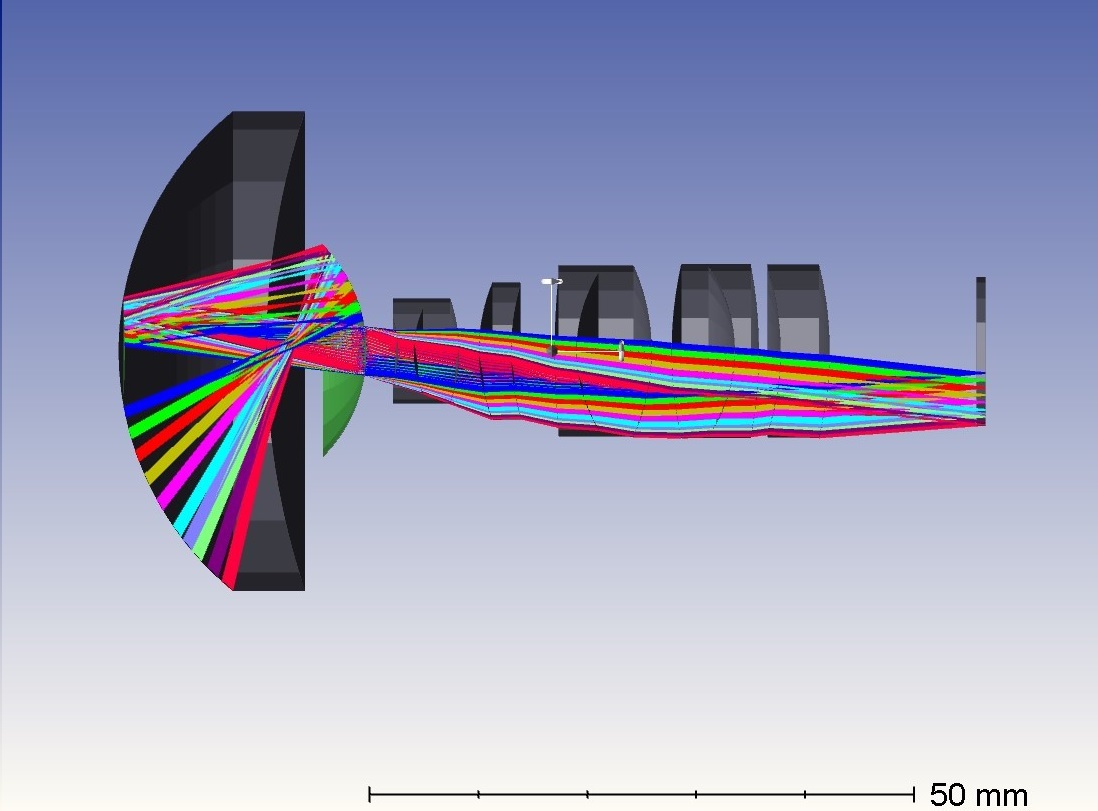}}
    \caption{Shaded model of the designed PAL imaging system. It consists of 6 groups of 10 lenses. Different colors represent light in different field of view. }
    \label{fig:1}
\end{figure}

The PAL system designed in this work is composed of $10$ standard spherical lenses, and the glass materials are all from the CDGM company.
The PAL block is composed of two glass materials glued together.
By coating the transmission film and reflection film on its surface, the light path can be folded. Its structure diagram is shown in Fig.~\ref{fig:1}. 

\begin{figure}[!t]
    \centerline{\includegraphics[width = 0.45\textwidth]{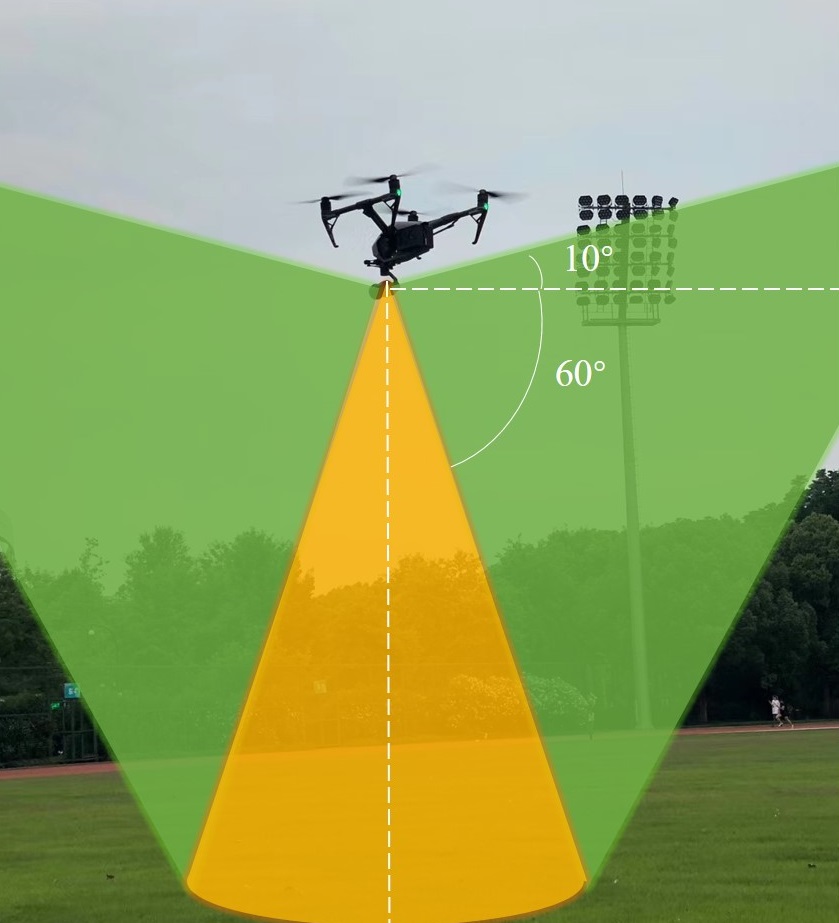}}
    \caption{The schematic diagram of the drone system. The green area represents the imaging area and the FoV is $(30^\circ\sim100^\circ)\times360^\circ$. The yellow area represents the blind area.}
    \label{fig:2}
\end{figure}

We have chosen the Inspire 2 series drone module developed by DJI to be equipped with a PAL, which is placed vertically downward to cover a wider field of view on the ground and to avoid the stray light problem caused by direct sunlight. The drone system is flying around $100m$ above the track and field to collect image data from multiple directions. The schematic diagram of the experiment is shown in Fig.~\ref{fig:2}. The lateral field of view of the PAL system involved in imaging is $10^\circ$ horizontally upward and $60^\circ$ horizontally downward, and the overall field of view is $360^\circ\times70^\circ$.

\subsection{Aerial Scene Segmentation Model}
 
For our segmentation method, there are three requirements. The model must be light enough to meet the real-time inference speed demand for future migration to portable computing devices. In addition, The model needs to parse the image with a high accuracy. Further, the model should have multi-scale receptive fields for panoramic images with a ultra-wide angle. Inspired by SwfitNet~\cite{swiftnet} and RFNet~\cite{rfnet}, we have designed a lightweight novel U-Net like model with multi-scale receptive fields.

\subsubsection{Model Architecture}

\begin{figure}[h]
    \centerline{\includegraphics[width = 0.485\textwidth]{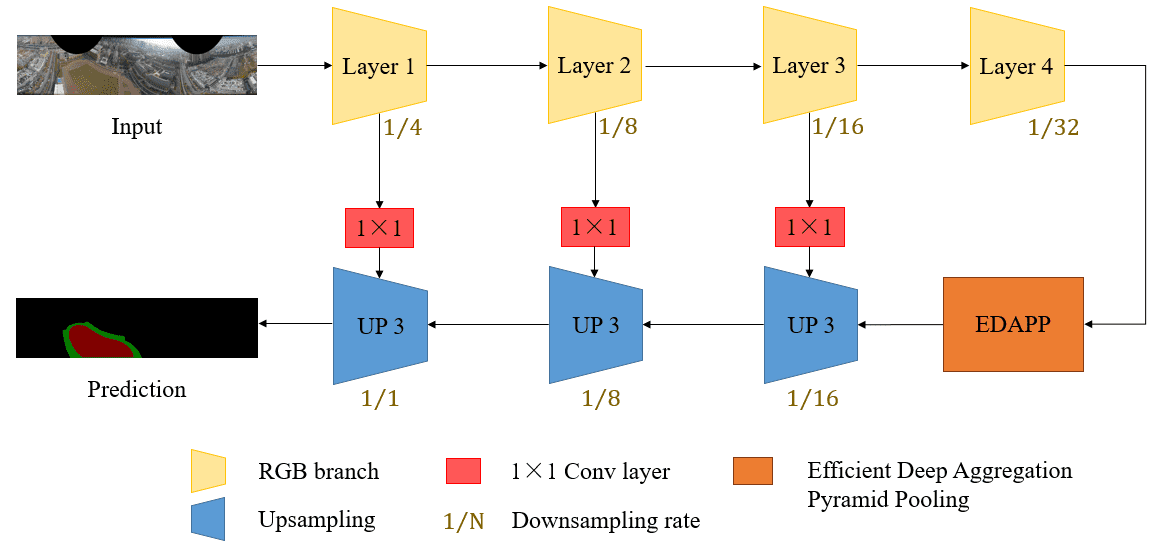}}
    \caption{The architecture of the proposed segmentation model.}
    \label{fig:model}
\end{figure}

The proposed network architecture is shown in Fig.~\ref{fig:model}.
We adopt ResNet-18~\cite{resnet} as our backbone, which is a mainstream light-weight feature extraction network.
With ImageNet~\cite{russakovsky2015imagenet} pre-trained weights, we can benefit from regularization induced by transfer learning and small operation footprint for real-time prediction.
The feature maps after each layer of ResNet is fused with the feature maps in the upsampling modules through skip connection with $1\times1$ convolution modules.
To increase the receptive field, the feature maps are transmitted to Efficient Deep Aggregation Pyramid Pooling module.
In the decoder, feature maps are added with lateral features from an earlier layer of the encoder element-wisely, and after convolution, the blended feature maps are upsampled by bilinear interpolation. Through skip connections, high-resolution feature maps full of detail information are blended with low-resolution feature maps with rich semantic information.

\subsubsection{Efficient Deep Aggregation Pyramid Pooling}
\begin{figure}[!t]
    \centerline{\includegraphics[width = 0.485\textwidth]{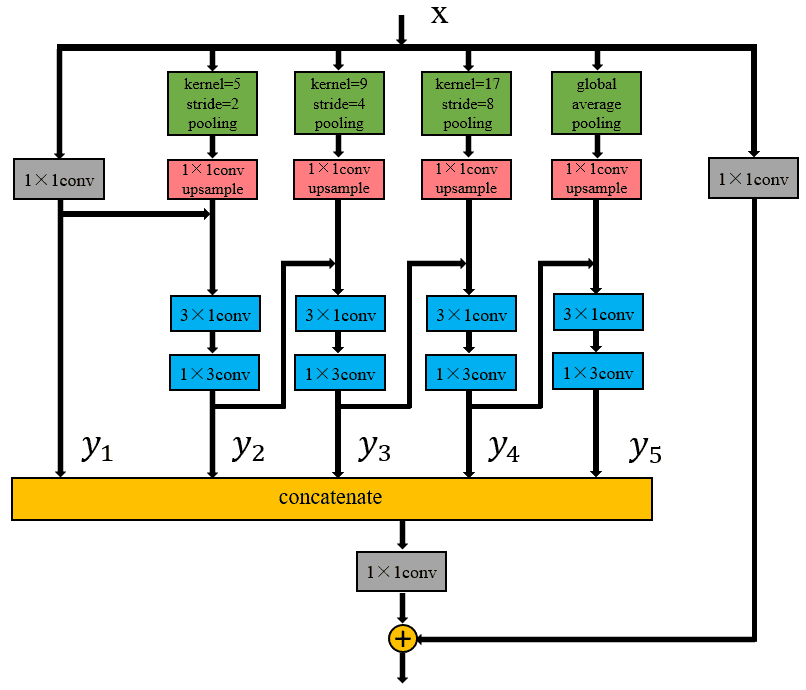}}
    \caption{The architecture of the EDAPP module.}
    \label{fig:edapp}
\end{figure}

For aerial panoramic images, many objects are rather small and only take up little ratio of pixels. Receptive field is extremely important in the task for a fine-grain segmentation.
Inspired by the context extraction module in DDRNet\cite{ddrnet}, we develop an Efficient Deep Aggregation Pyramid Pooling (EDAPP) module.
Fig.~\ref{fig:edapp} shows the architecture of EDAPP.
First, we perform average pooling with kernel size of $5$, $9$, $17$, and global pooling respectively.
Single $3\times3$ or $1\times1$ convolutions in Spatial Pyramid Pooling (SPP)~\cite{SPP} is not enough, so after $1\times1$ convolution upsampling to the same resolution, to efficiently blend the multi-scale contextual information better, we propose to leverage a combination of $3\times1$ convolution and $1\times3$ convolution. Another stream consists of only a $1\times1$ convolution. Asserting an input $x$, the process can be summarized in the following:

\begin{align}
\footnotesize
    y_{k} =\left\{ \begin{array}{l}
    C_{1\times1}(x), \hfill k=1;  \\
    C_{1\times3}(C_{3\times1}(UP(C_{1\times1}(P_{2^{k}+1,2^{k-1}}(x)))+y_{k-1}), \hfill 1<k<n;  \\
    C_{3\times3}(UP(C_{1\times1}(P{global}(x)))+y_{k-1}), \hfill k=n. \\
 \end{array} \right.
\label{edapp_equotion}
\end{align}

Here, $C$ denotes convolution, $UP$ denotes bilinear upsample, $P$ and $P_{global}$ denote pooling and global pooling respectively.
$i$ and $j$ of the $P_{i,j}$ denote the kernel size and stride of the pooling layer. 
A pair of $3\times1$ convolution and $1\times3$ convolution have the same receptive field as a single $3\times3$ kernel, enhanced directional feature extraction, but less computational complexity and faster inference speed.
This architecture helps extracting and integrating deep information with different scales by different pooling kernel sizes. All the features maps are concatenated and blended through a $1\times1$ convolution. Finally, a $1\times1$ convolution compress all the feature maps and after that we also add a skip connection with a $1\times1$ convolution for easier optimization.

\section{Experiments}

\subsection{Aerial-PASS Dataset}

We collected data in 4 different places using the UAV with a PAL camera, and all the data were under sufficient illumination conditions. In the height of about $100$ meters, we collected videos with the length of about $3$ hours. Limited to the size of the CMOS in the camera, the image produced can't show all the imaging plane of the lens.
In the following deployment phase, the PAL system was calibrated using the interface provided by the omnidirectional camera toolbox~\cite{calitoolbox}. Before training, the PAL image was unfolded to a familiar rectangle image. The unfolded process is depicted in the following equations:

$$
i = \frac{r-r_{1}}{r_{2}-r_{1}} \times{height}
\eqno{(2)}
\label{unfold_equation_1}
$$

$$
j = \frac{\theta}{2\pi} \times{width}
\eqno{(3)}
\label{unfold_equation_2}
$$

\begin{figure}[!t]
    \centerline{\includegraphics[width = 0.485\textwidth]{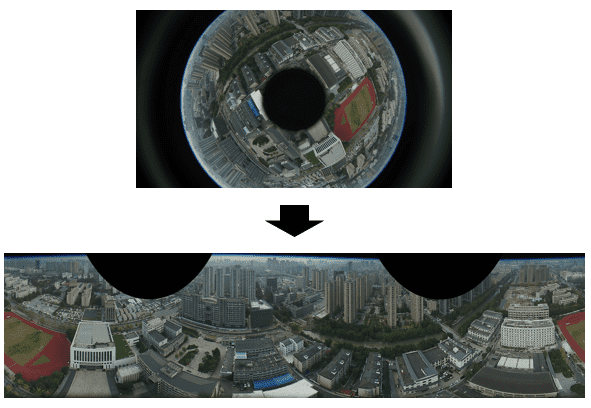}}
    \caption{The unfolding process of the PAL image. Limited to the size of CMOS, top and bottom sides of the imaging plane are blocked in the image, resulting in two scalloped shadows in the unfolded image.}
    \label{fig:PAL_unfolding}
\end{figure}

Here, $i$ and $j$ denote the index of x and y axis of the unfold image, respectively. $r_{1}$ and $r_{2}$ are the internal and external radii of the PAL image. Width and height are the image size of the unfolded image.
In our experiment, we unfolded the PAL image to a $2048\times512$ image. Fig.~\ref{fig:PAL_unfolding} shows the unfolding process.

We annotated all $462$ images out of the $3$-hour-long video.
We created pixel-wise fine labels on the most critical classes relevant to the application of track detection, and we randomly split out $42$ images for the test set. As far as we know, This is the first aerial panoramic image dataset with semantic segmentation annotations in the community.

\subsection{Training Details}

All the experiments were implemented with PyTorch 1.3 on a single 2080Ti GPU with CUDA10.0, cuDNN 7.6.0.
We chose Adam~\cite{kingma2014adam} for optimization with a learning rate of $5\times10^{-4}$, where cosine annealing learning rate scheduling policy~\cite{loshchilov2016sgdr} is adopted to adjust the learning rate with a minimum value of $5\times10^{-4}$ in the last epoch and weight decay was set to $1\times10^{-4}$.
The ResNet-18~\cite{resnet} backbone was initialized with pre-trained weights from ImageNet~\cite{russakovsky2015imagenet} and the rest part of the model was initialized with the Kaiming initialization method~\cite{kaiminginitialization}.
We updated the pre-trained parameters with $4$ times smaller learning rate and weight decay rate.
The data augment operations consist of scaling with random factors between $0.5$ and $2$, random horizontal flipping, and random cropping with an output resolution of  $512\times512$.
Models were trained for 100 epochs with a batch-size of $6$. We used the standard ``Intersection over Union (IoU)'' metric for the evaluation:
$$
    IoU = \frac{TP}{TP+FP+FN}
    \label{Iouacc}
    \eqno{(4)}
$$

\subsection{Results and Analysis}

Based on the Aerial-PASS dataset, we have created a benchmark to compare our proposed network with other two competitive real-time semantic segmentation networks: the single-scale SwiftNet~\cite{swiftnet} (similar backbone with our network) and ERF-PSPNet~\cite{erf_pspnet} (a lightweight network designed for panoramic segmentation~\cite{pass}).
All the networks were trained on the training set of Aerial-PASS with the same training strategy and tested on the testing set of the dataset. Table~\ref{pal-accuracy} shows the numerical performance comparisons of the three efficient networks. Our proposed model outperforms both networks designed for panoramic segmentation (ERF-PSPNet) and semantic segmentation (SwifNet) by clear gaps. 

\begin{table}[!t]
\caption{Per-class and mean IoU of our method and other two networks on the testing set of Aerial-PASS dataset.}
\label{pal-accuracy}
\begin{tabular}{l|l|l|l|l}
\textbf{Network}      & \textbf{Track} & \textbf{Field} & \textbf{Others} & \textbf{Mean} \\ \hline
ERF-PSPNet            & 64.16\%        & 97.67\%        & 92.02\%         & 84.62\%       \\ \hline
SwiftNet (single scale) & 63.15\%        & 98.76\%        & 91.63\%         & 84.52\%       \\ \hline
\textbf{Ours}         & 67.67\%        & 99.06\%        & 92.16\%         & 86.30\%      
\end{tabular}
\end{table}

\begin{figure*}
    \centering
    \includegraphics[width=1\linewidth]{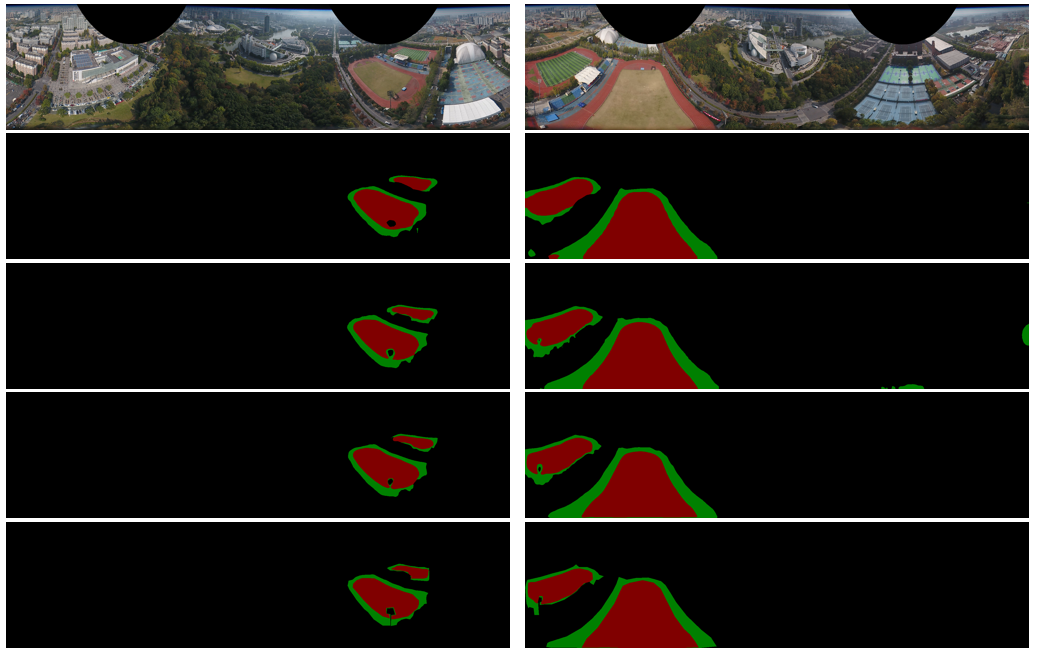}
    \caption{Qualitative semantic segmentation results. From top to bottom row: RGB input image, ERF-PSPNet, SwiftNet, our method, and ground truth.}
    \label{fig:contrast}
\end{figure*}

Fig.~\ref{fig:contrast} shows visualizations of inference labels of our proposed method and other two models, in which green denotes the track and red denotes the field. All the input images are the unfolded PAL images.
The labels show the qualitative result of the proposed method.
As we can find that our method performs well in both large-scale objects like field and small-scales objects like the boundary of track and other objects thanks to our EDAPP module designed for multi-scale feature learning.

\subsection{Comparison with the State of the Art}

\begin{figure}[ht]
    \centering
    \subfigure[RGB]{
    \begin{minipage}[b]{0.32\linewidth}
    \includegraphics[width=1\linewidth]{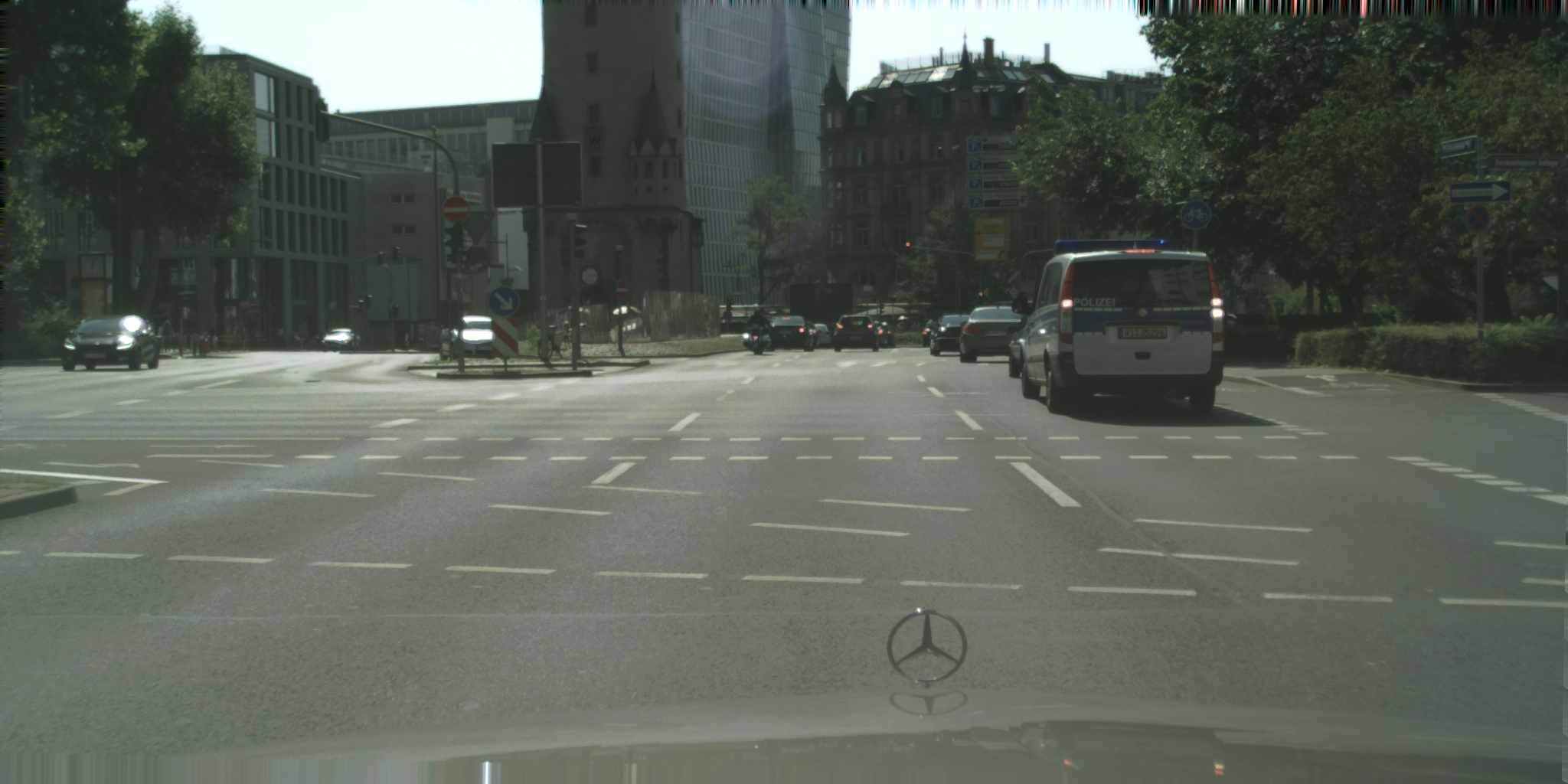}\vspace{0pt}  
    \includegraphics[width=1\linewidth]{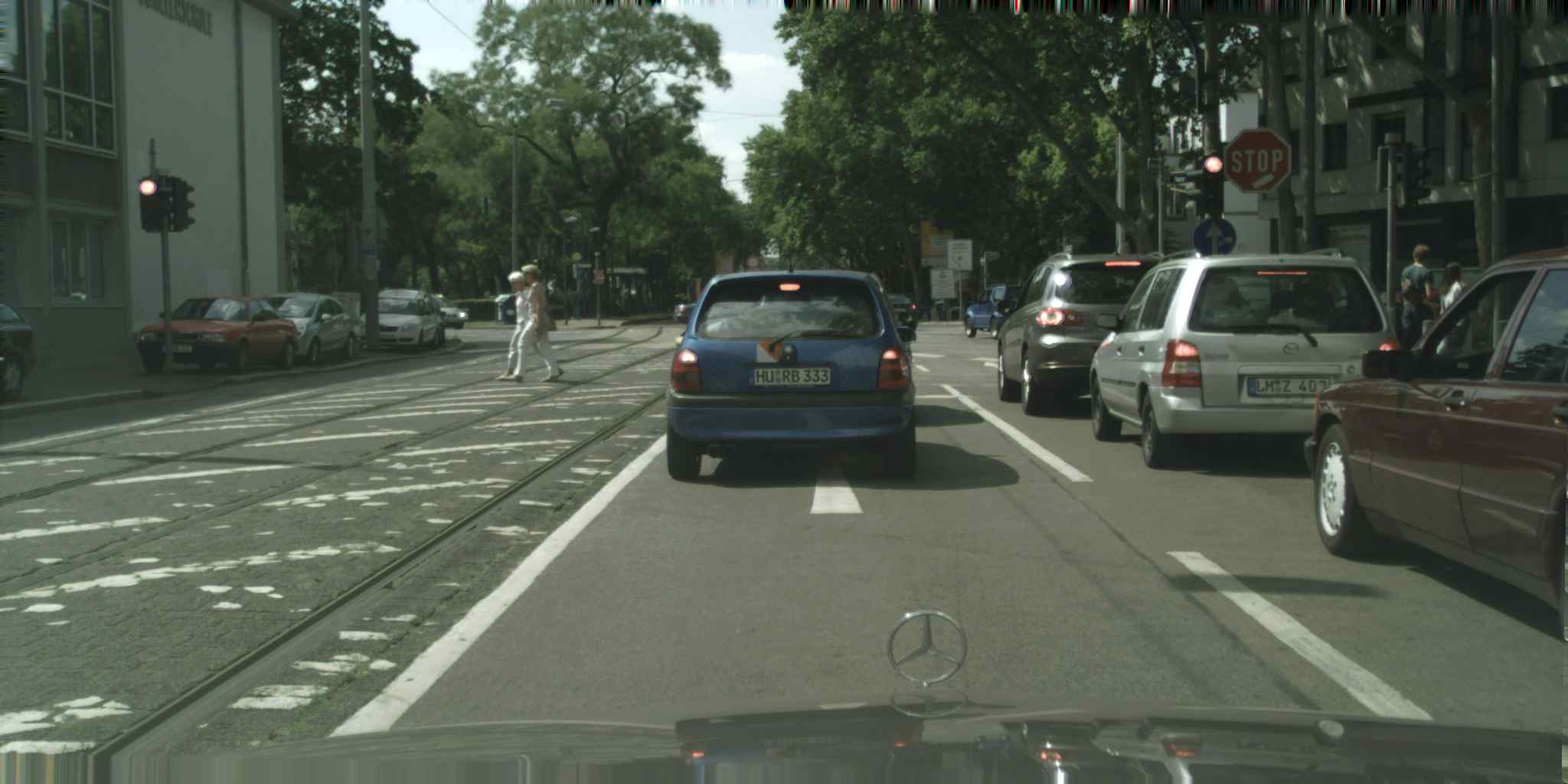}\vspace{0pt}
    \includegraphics[width=1\linewidth]{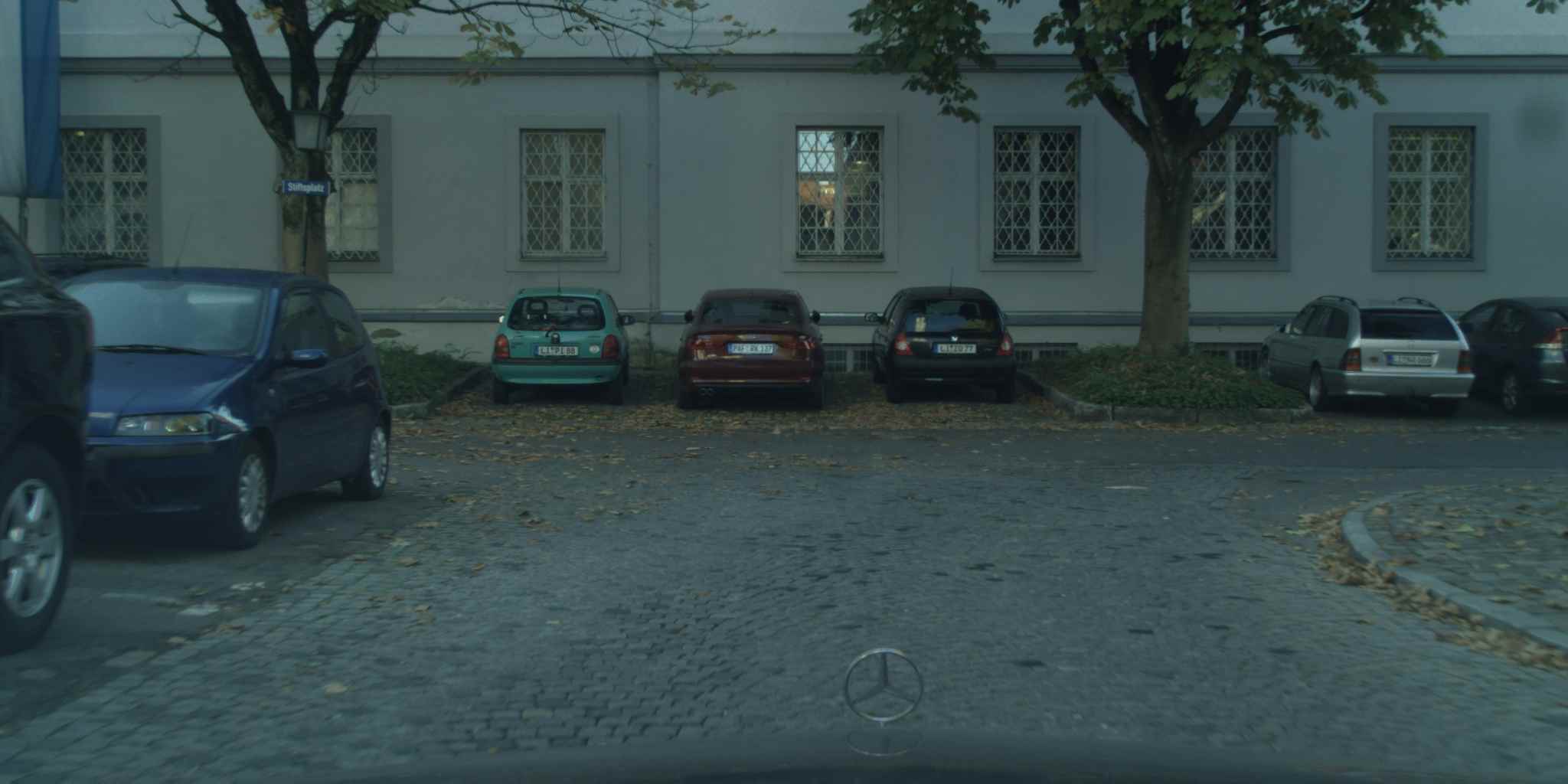}\vspace{0pt}
    \includegraphics[width=1\linewidth]{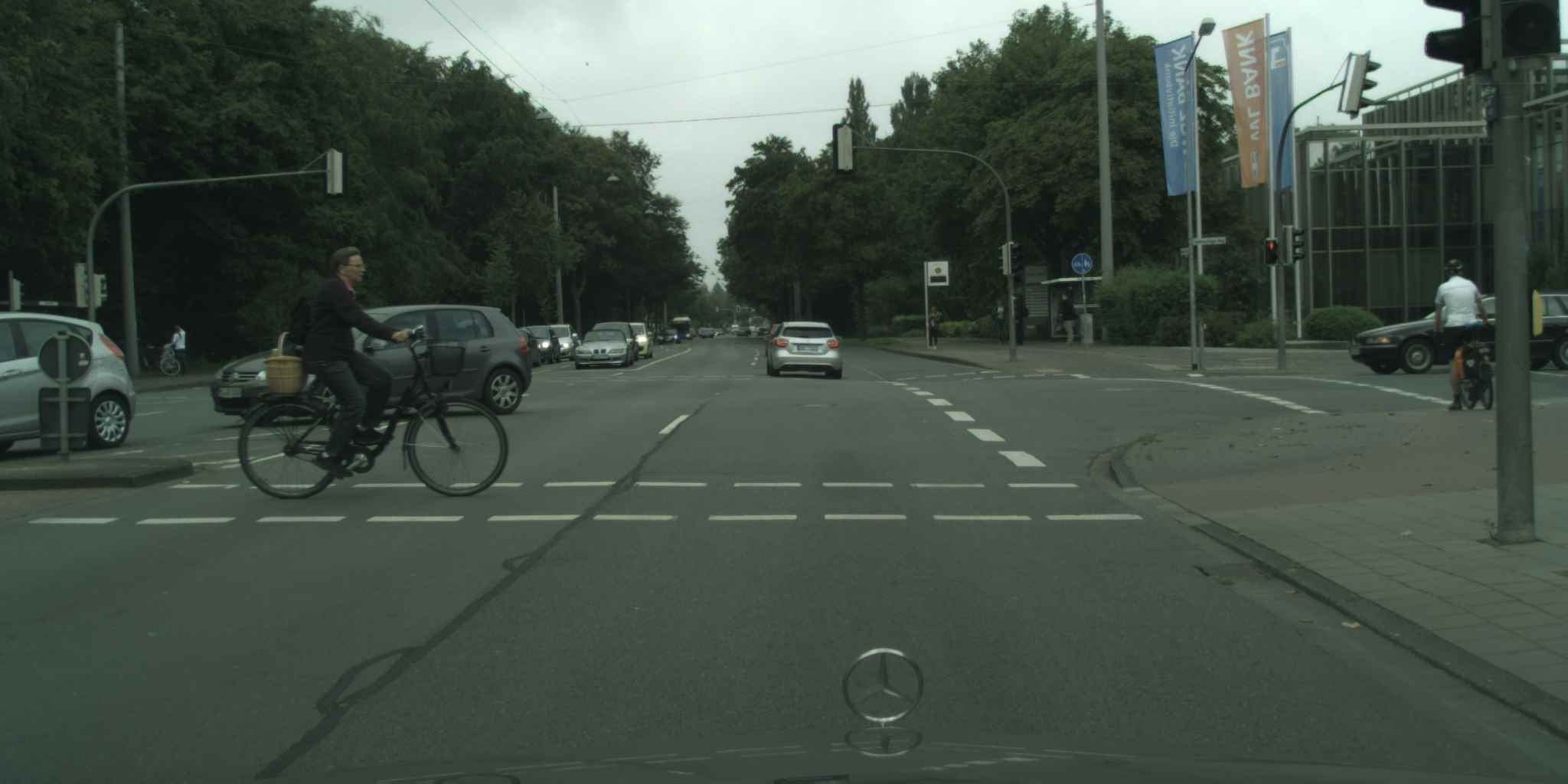}\vspace{0pt}
    \includegraphics[width=1\linewidth]{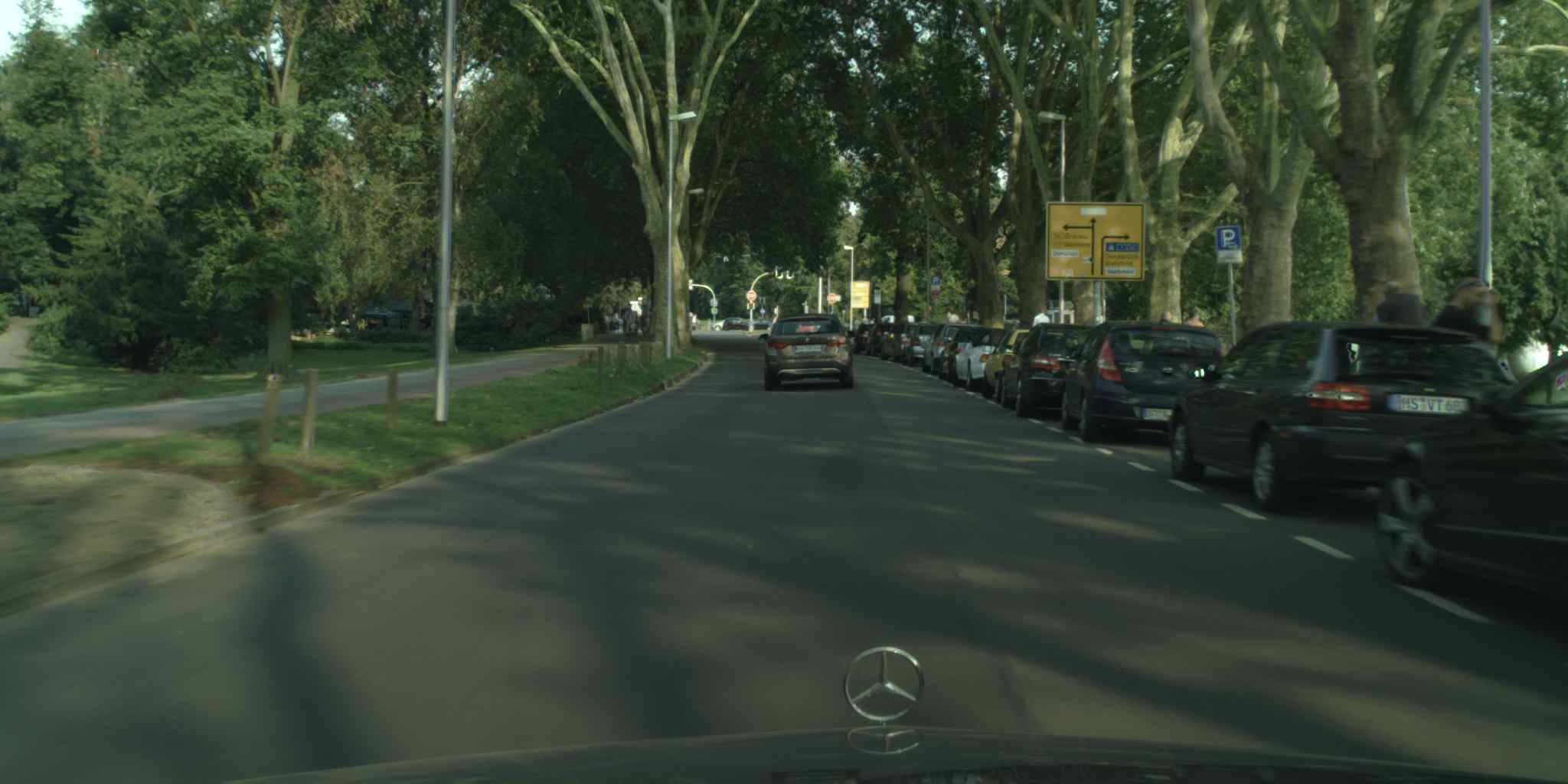}
    \end{minipage}}\hspace{-6pt}  
    \subfigure[GT]{
    \begin{minipage}[b]{0.32\linewidth}
    \includegraphics[width=1\linewidth]{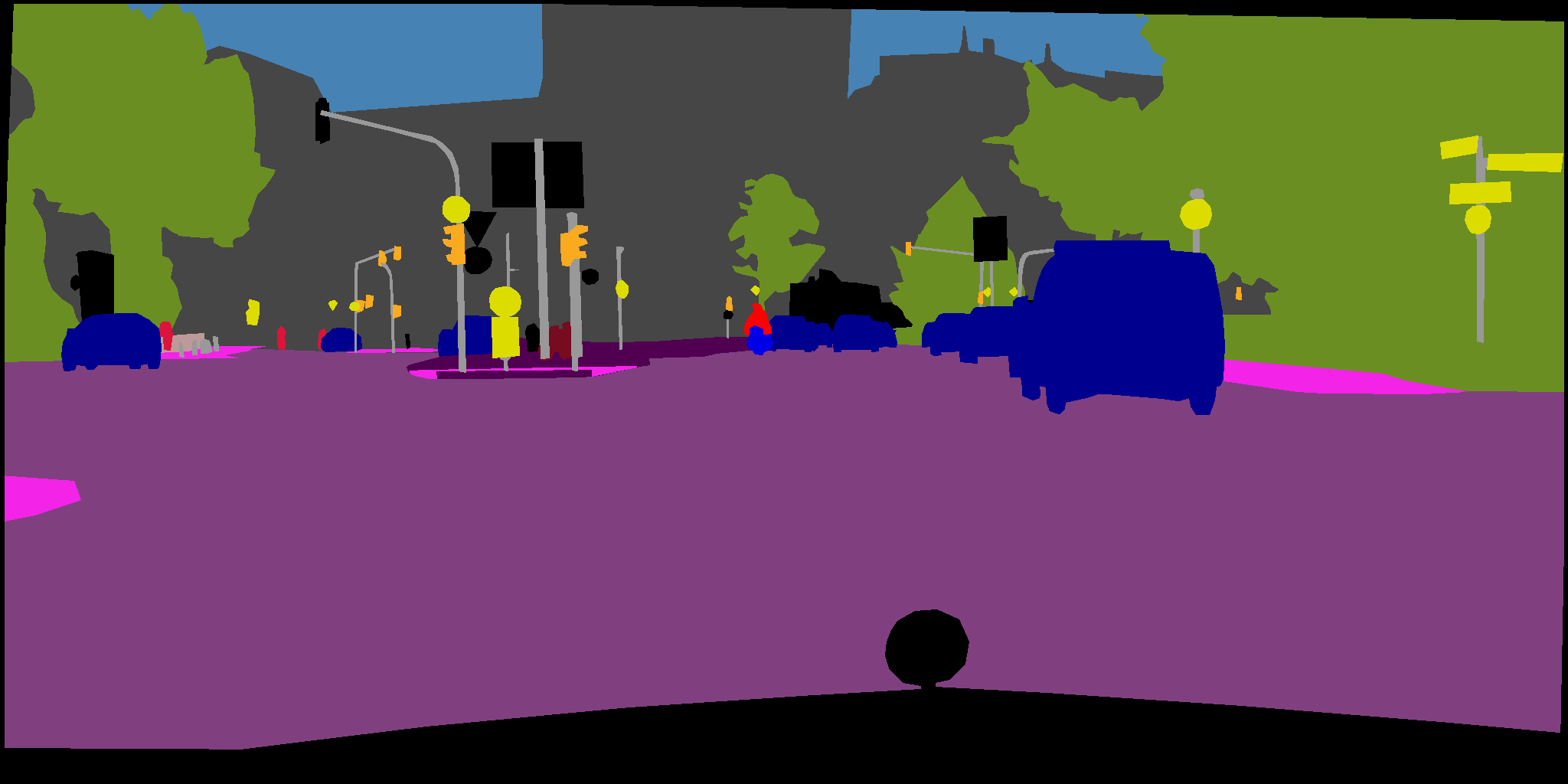}\vspace{0pt}  
    \includegraphics[width=1\linewidth]{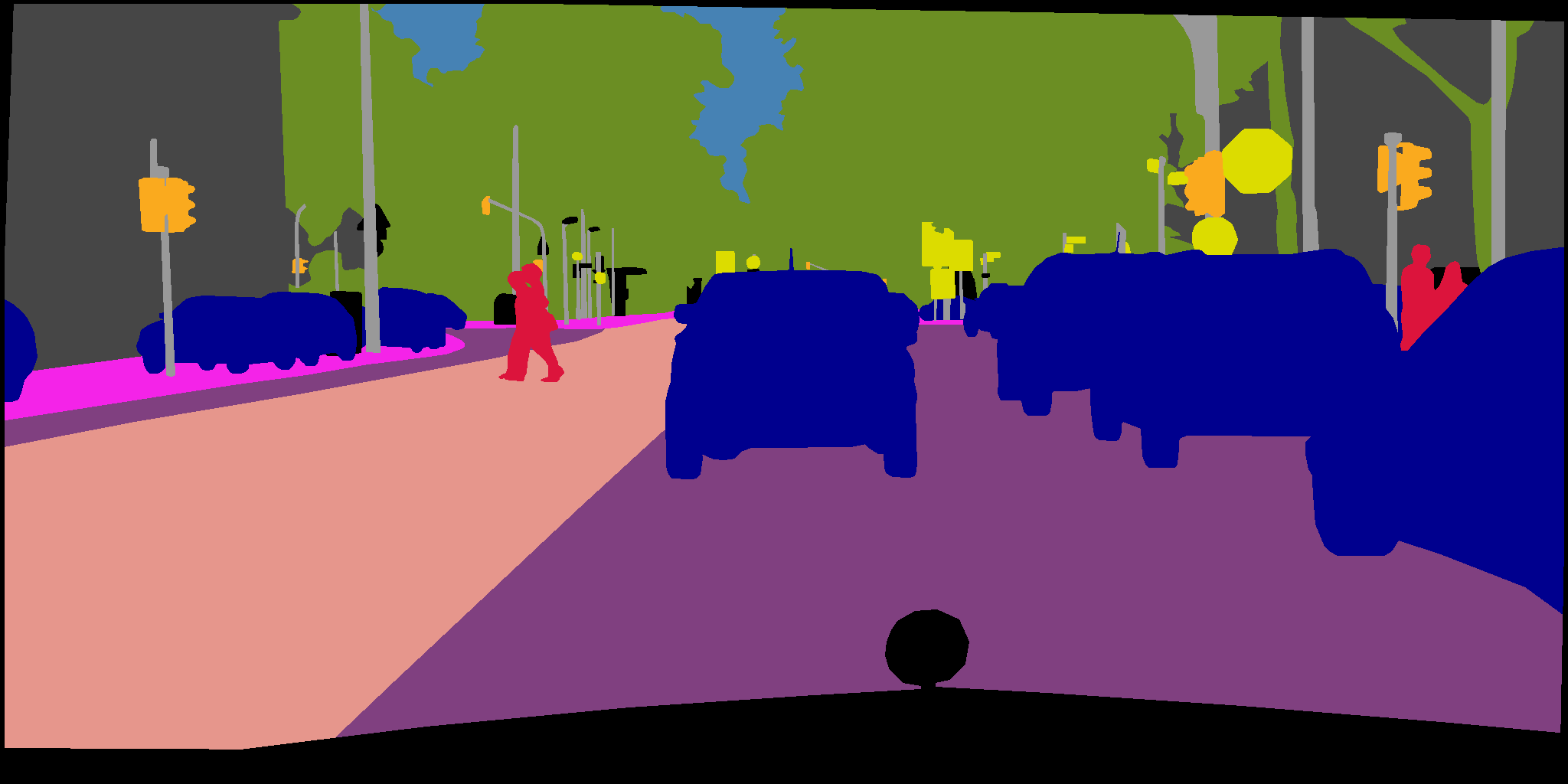}\vspace{0pt}  
    \includegraphics[width=1\linewidth]{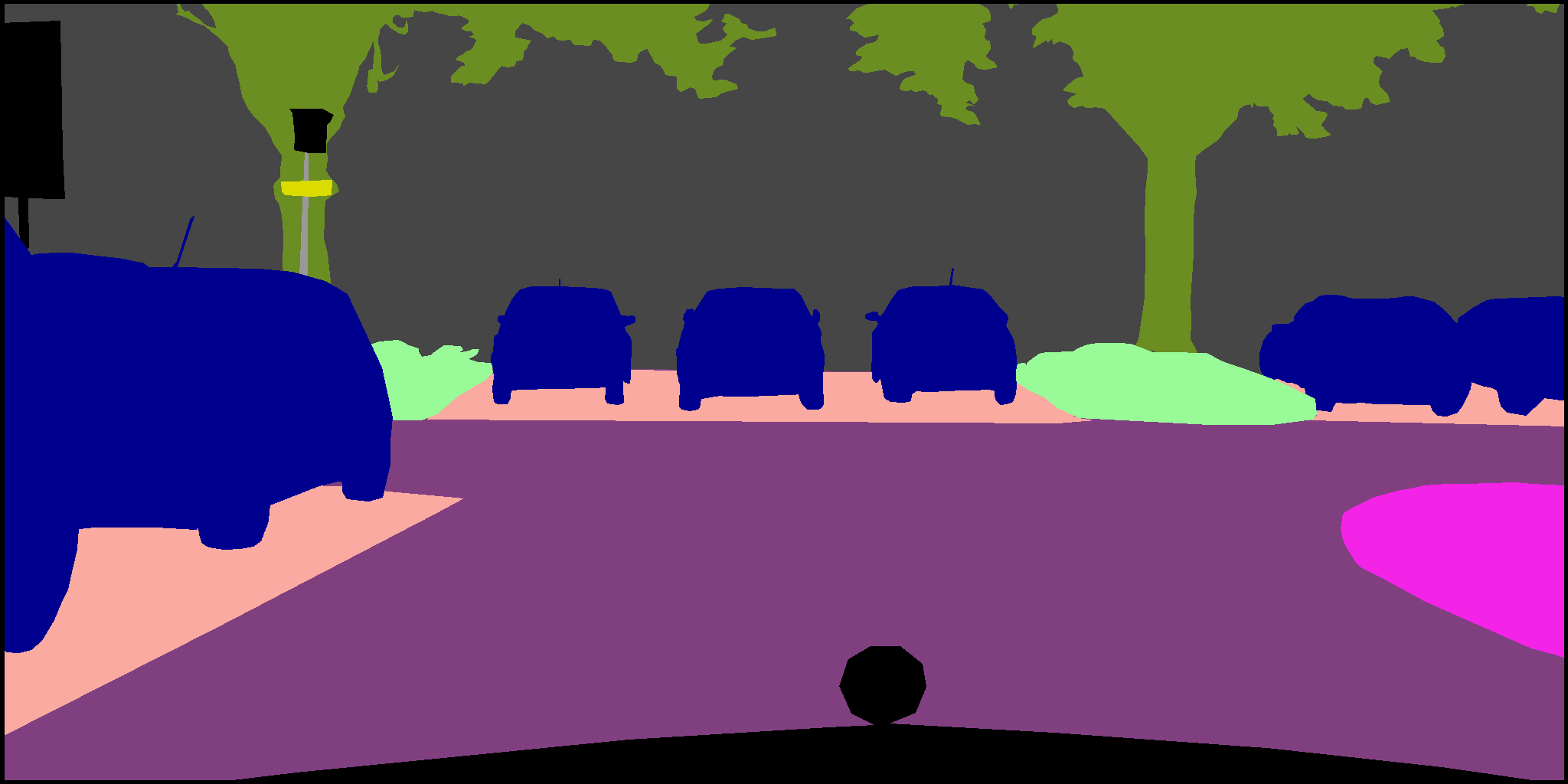}\vspace{0pt}  
    \includegraphics[width=1\linewidth]{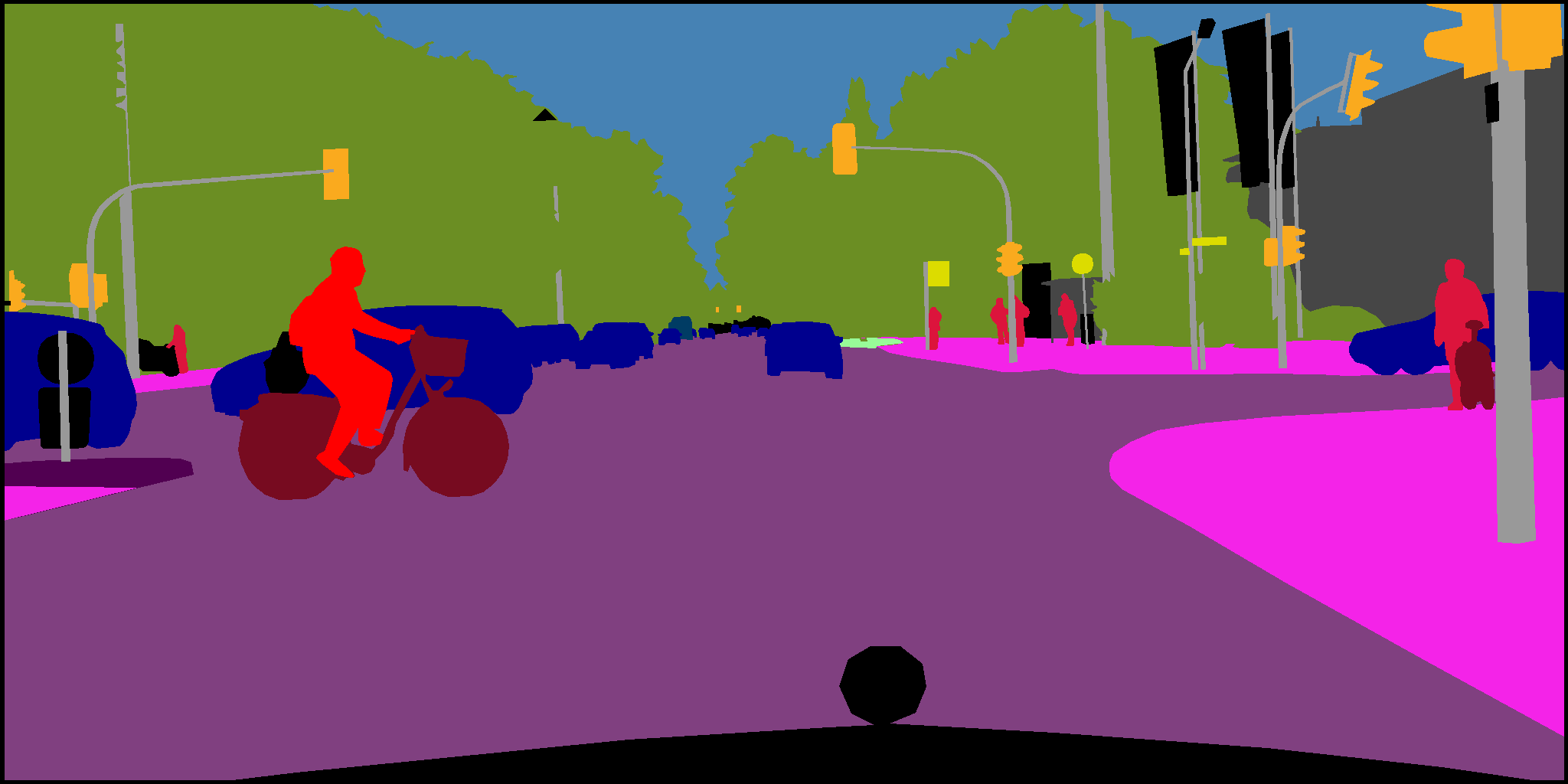}\vspace{0pt}  
    \includegraphics[width=1\linewidth]{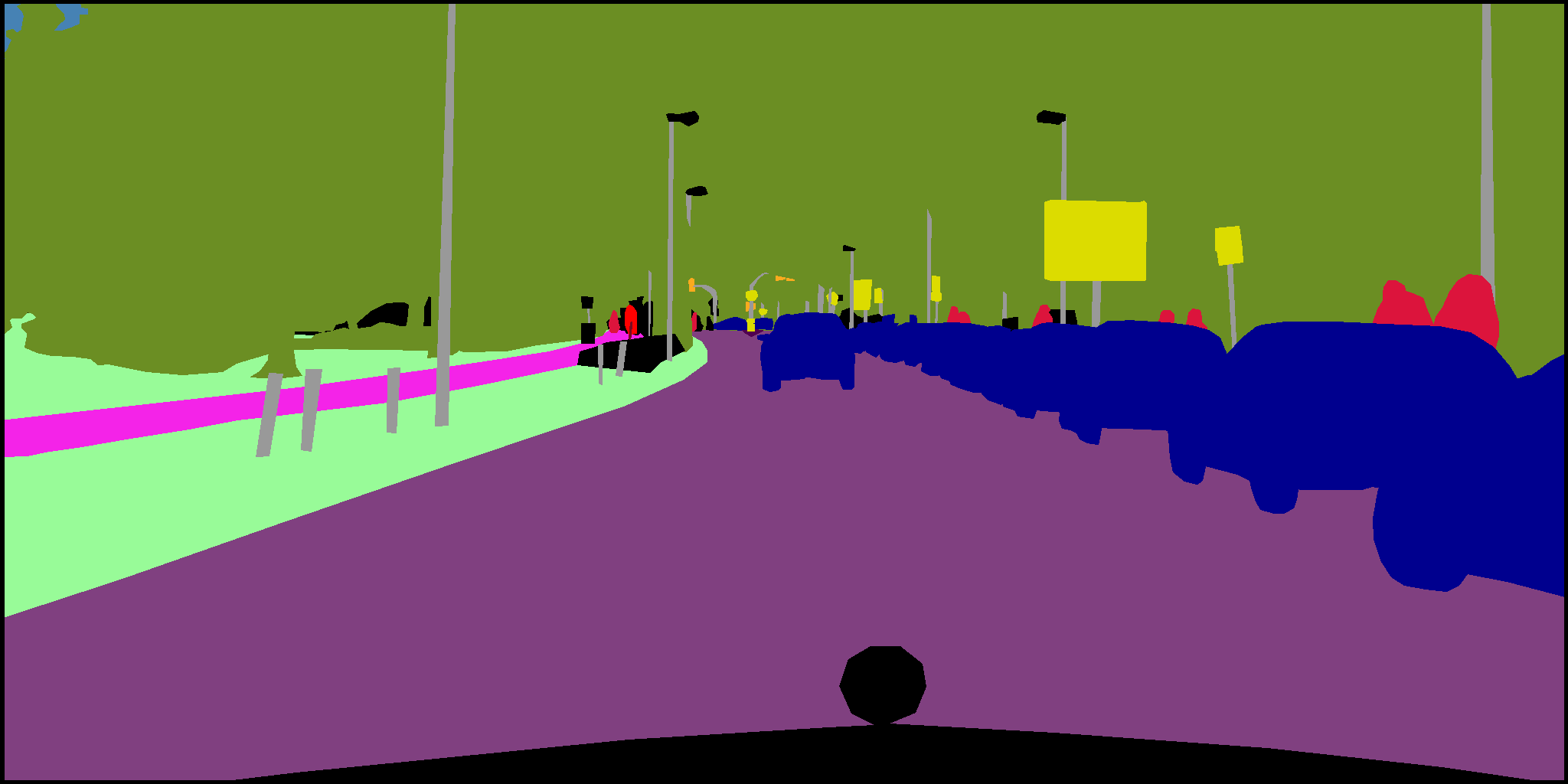}
    \end{minipage}}\hspace{-6pt}
    \subfigure[Result]{
    \begin{minipage}[b]{0.32\linewidth}
    \includegraphics[width=1\linewidth]{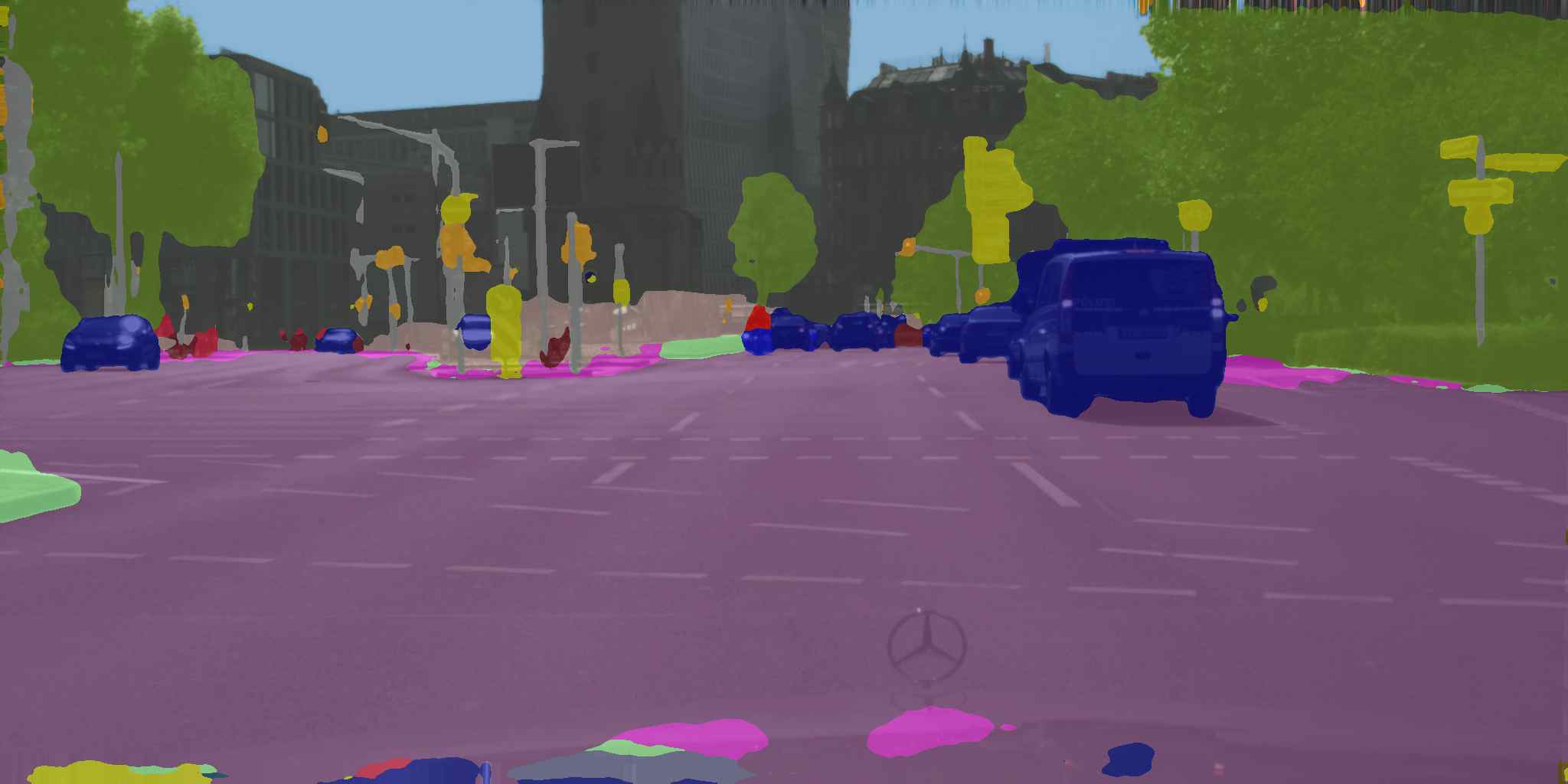}\vspace{0pt}  
    \includegraphics[width=1\linewidth]{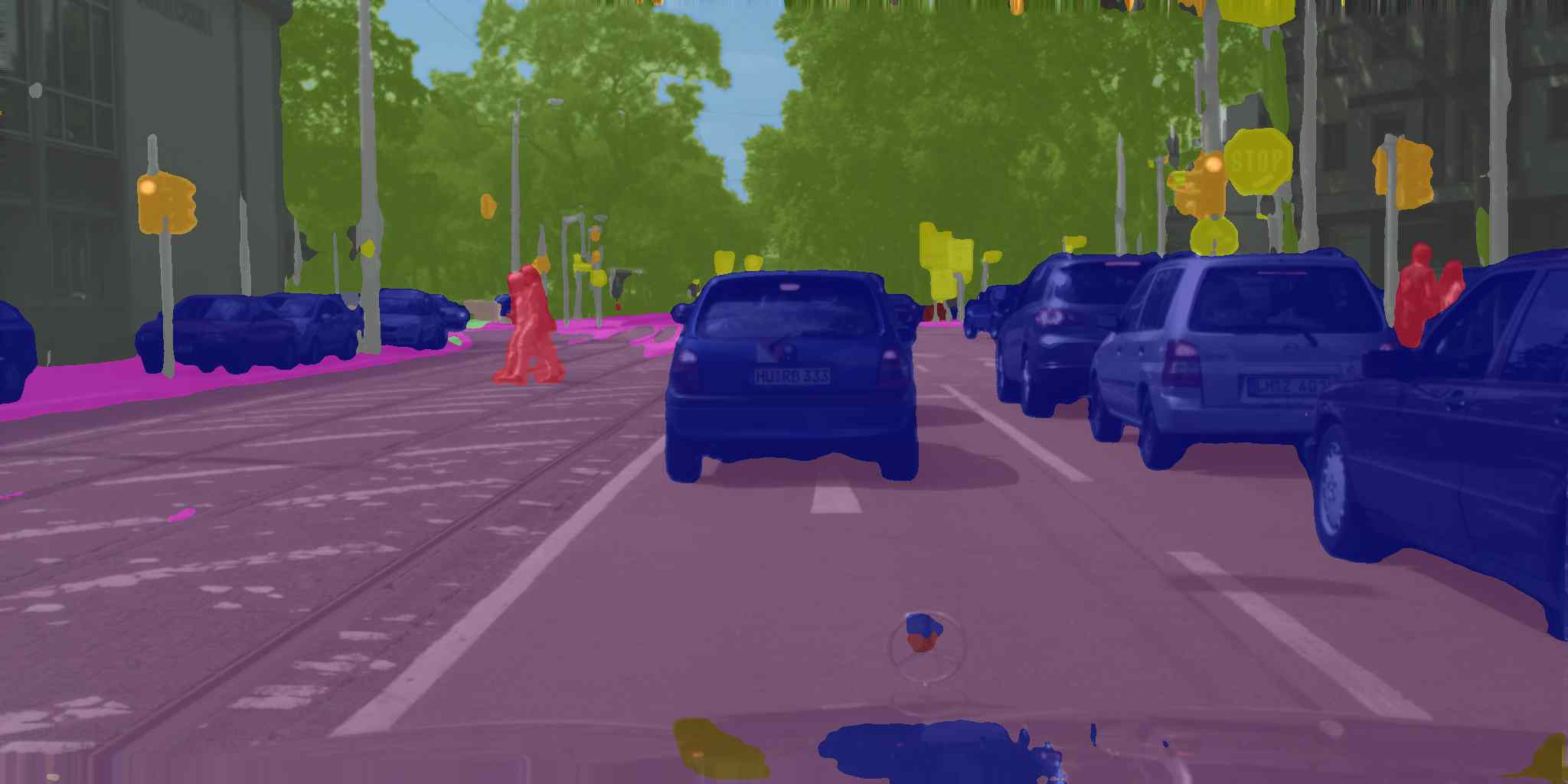}\vspace{0pt}
    \includegraphics[width=1\linewidth]{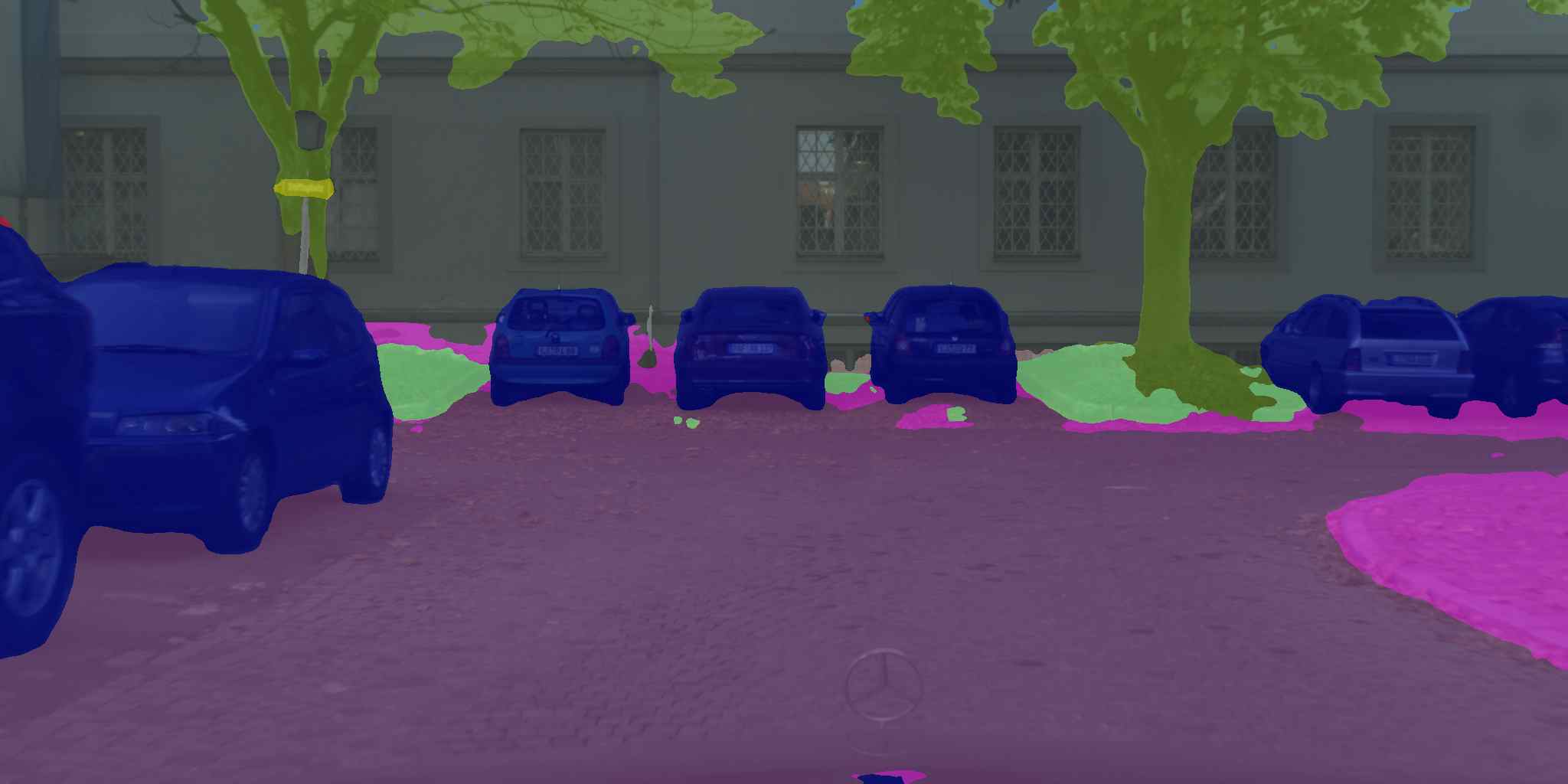}\vspace{0pt}
    \includegraphics[width=1\linewidth]{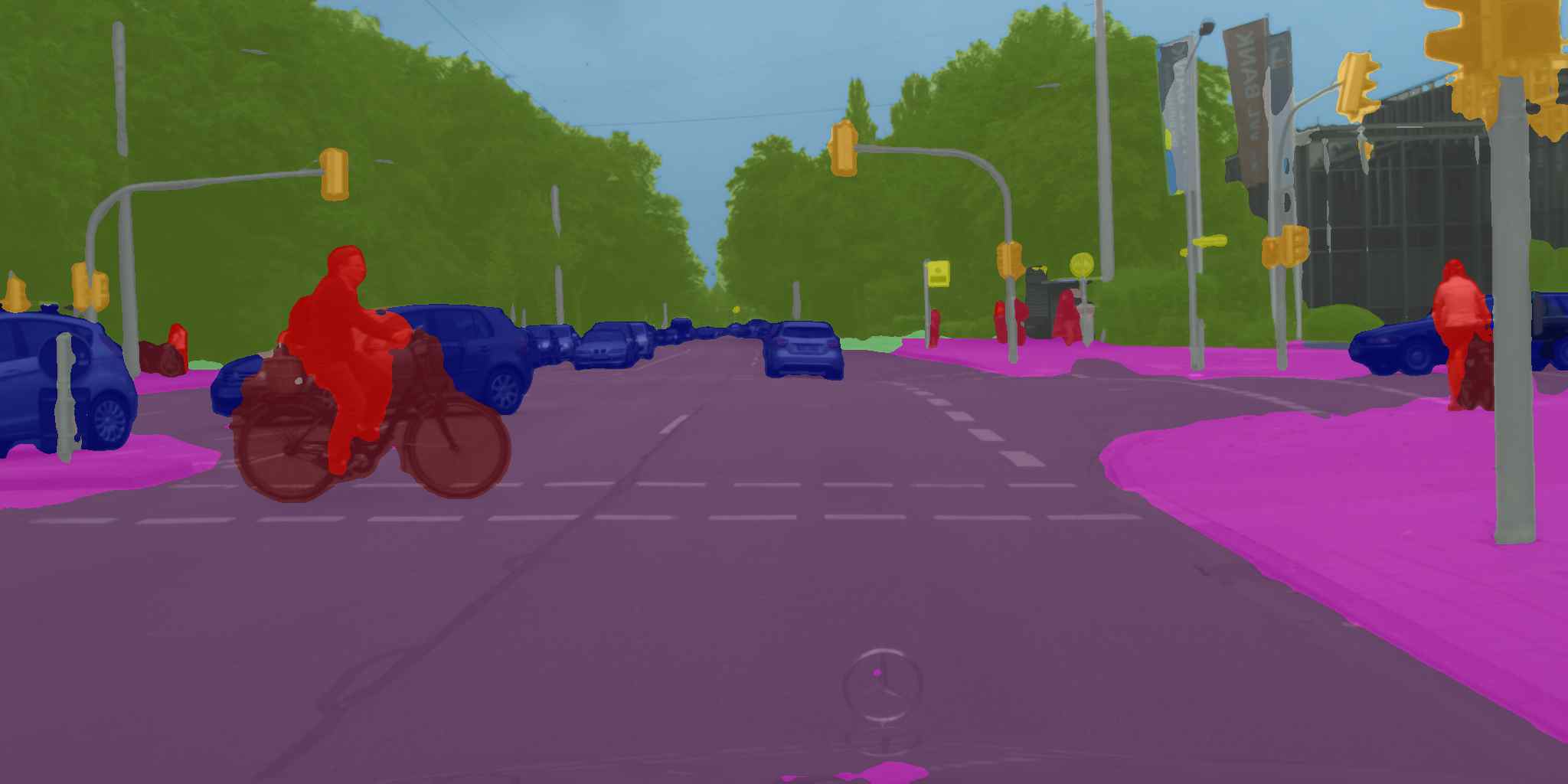}\vspace{0pt}
    \includegraphics[width=1\linewidth]{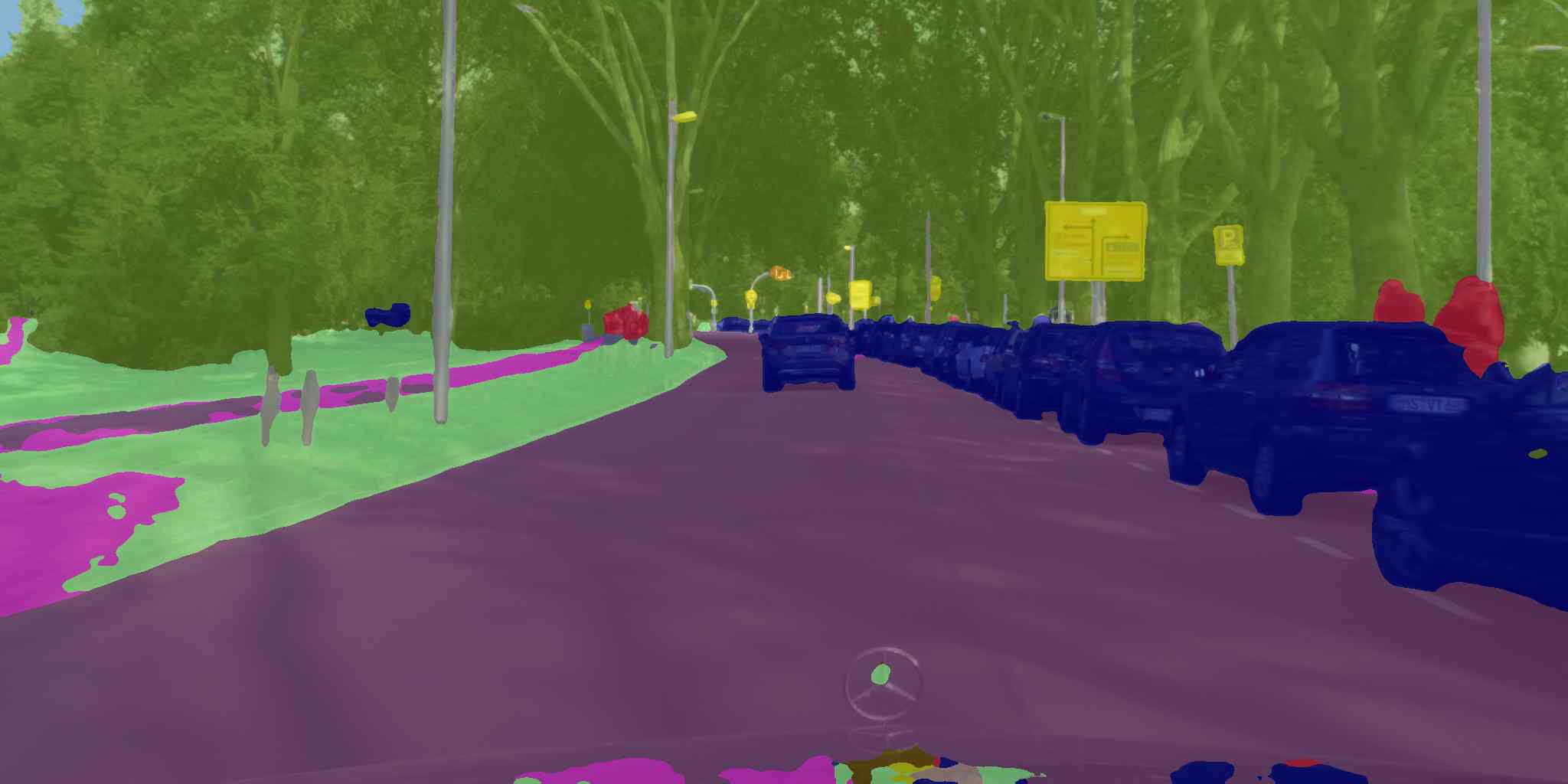}
    \end{minipage}}\hspace{-6pt}
    \caption{Qualitative results of our method on the \emph{Cityscapes} dataset.}
    \label{fig:cityscapes}
\end{figure}

To further compare with other state-of-the-art network, we also trained our network on \emph{Cityscapes}~\cite{Cordts2016Cityscapes}, which is a large-scale RGB dataset that focuses on semantic understanding of urban street scenes.
It contains $2975$/$500$/$1525$ images in the training/validation/testing subsets, both with finely annotated labels on $19$ classes.
The images cover $50$ different cities with a full resolution of $2048\times1024$.
We trained our network on the training set of the \emph{Cityscapes} dataset and test our network on the validation set. Table~\ref{table:compare} shows the IoU result of our method and other mainstream real-time semantic segmentation models.
Our method has achieved an excellent balance between accuracy and inference speed. Fig.~\ref{fig:cityscapes} shows some representative inference results of our proposed method. Overall, the qualitative results verify the generalization capacity of our proposed network for both challenging large-FoV aerial image segmentation and high-resolution driving scene segmentation. 

\begin{table}[htbp]
\centering
\caption{Comparison of semantic segmentation methods on the validation set of Cityscapes.}
\label{table:compare}
\begin{tabular}{llll}
\multicolumn{1}{l|}{\textbf{Network}} & \multicolumn{1}{l|}{\textbf{MIoU}} &  \multicolumn{1}{l}{\textbf{Speed (FPS)}}  &  \\ \cline{1-3}
\multicolumn{1}{l|}{FCN8s~\cite{fcn}}            & \multicolumn{1}{l|}{65.3\%}        &     \multicolumn{1}{c}{2.0$^{\mathrm{*}}$}                &  \\
\multicolumn{1}{l|}{DeepLabV2-CRF~\cite{deeplab}}    & \multicolumn{1}{l|}{70.4\%}        & \multicolumn{1}{c}{n/a}                 &  \\
\multicolumn{1}{l|}{ENet~\cite{paszke2016enet}}             & \multicolumn{1}{l|}{58.3\%}        & \multicolumn{1}{c}{76.9$^{\mathrm{*}}$}            &  \\
\multicolumn{1}{l|}{ERF-PSPNet~\cite{erf_pspnet}}       & \multicolumn{1}{l|}{64.1\%}        & \multicolumn{1}{c}{20.4}                &  \\
\multicolumn{1}{l|}{SwiftNet~\cite{swiftnet}}         & \multicolumn{1}{l|}{72\%}          & \multicolumn{1}{c}{41.0}                &  \\ \cline{1-3}
\multicolumn{1}{l|}{Ours}             & \multicolumn{1}{l|}{72.8\%}        & \multicolumn{1}{c}{39.4}                &  \\
\multicolumn{3}{l}{$^{\mathrm{*}}$ Speed on half resolution images.}
\end{tabular}
\end{table}

\section{Conclusion}

In this study, we propose a lightweight UAV system Aerial-PASS with a designed Panoramic Annular Lens (PAL) camera and a real-time semantic segmentation network for aerial panoramic image collection and segmentation.
The minimization-dedicated PAL camera equipped in the UAV can be used for collecting annular panoramic images without requiring a complicated servo system to control the attitude of the lens.
To classify the track and field in the images at the pixel wise, we collect and annotate $462$ images and propose an efficient semantic segmentation network.
The proposed network has multi-scale reception fields and an efficient backbone, which outperforms other competitive networks on our Aerial-PASS dataset and also has reached the state-of-the-art performance on the \emph{Cityscapes} dataset with 39.4 Hz in full resolution on a single 2080Ti GPU processor.
In the future, we aim to transplant the algorithm to the portable embedded GPU on the UAV for more field tests.

\bibliographystyle{ieeetr}
\bibliography{bib}

\end{document}